\documentclass{article}

\usepackage[numbers]{natbib}

\usepackage[final]{neurips_2024}

\usepackage{graphicx}
\usepackage[utf8]{inputenc} 
\usepackage[T1]{fontenc}    
\usepackage{hyperref}       
\usepackage{url}            
\usepackage{booktabs}       
\usepackage{amsfonts}       
\usepackage{nicefrac}       
\usepackage{microtype}      
\usepackage{xcolor}         
\usepackage{wrapfig}

\usepackage{hyperref}
\usepackage{url}
\usepackage{graphicx}
\usepackage{subcaption}

\usepackage{multirow, multicol}
\usepackage{booktabs}

\usepackage{amsmath}
\usepackage{amssymb}
\usepackage{mathtools}
\usepackage{amsthm}
\usepackage{svg}
\usepackage[capitalize,noabbrev]{cleveref}

\usepackage{cancel}
\usepackage{bm}

\title{Dynamic Vocabulary Pruning in Early-Exit LLMs}

\author{%
  Jort Vincenti$^{1,}$\thanks{Equal contributions. Corresponding author: $<$\texttt{m.jazbec@uva.nl}$>$} \quad Karim Abdel Sadek$^{1, 3,*}$ \quad Joan Velja$^{1,*}$ \quad \textbf{Matteo Nulli}$^{1,*}$ \\ \textbf{Metod Jazbec}$^{1, 2}$\\
  $^1$University of Amsterdam $^2$UvA-Bosch Delta Lab $^3$Krueger AI Safety Lab (KASL)
}

\begin{document}

\maketitle

\begin{abstract}
  Increasing the size of large language models (LLMs) has been shown to lead to better performance. However, this comes at the cost of slower and more expensive inference. Early-exiting is a promising approach for improving the efficiency of LLM inference by enabling next token prediction at intermediate layers. Yet, the large vocabulary size in modern LLMs makes the confidence estimation required for exit decisions computationally expensive, diminishing the efficiency gains. To address this, we propose dynamically pruning the vocabulary at test time for each token. Specifically, the vocabulary is pruned at one of the initial layers, and the smaller vocabulary is then used throughout the rest of the forward pass. Our experiments demonstrate that such post-hoc dynamic vocabulary pruning improves the efficiency of confidence estimation in early-exit LLMs while maintaining competitive performance.
\end{abstract}

\section{Introduction}
Large language models (LLMs) are increasingly being adopted due to their impressive performance and their few-shot ability to adapt to new tasks \cite{brown2020language}. However, their growing size results in slow and costly inference. This is particularly limiting in environments with constrained resources or low-latency requirements (e.g., on-device). The push for more efficient LLM implementations is further motivated by growing concerns over their carbon footprint \citep{lannelongue2021green}. As a result, making LLMs more efficient at test time has recently received a lot of attention \citep{bai2024beyond, zhou2024survey, xu2024survey, kim2023full}. One promising paradigm for more efficient inference is \emph{early-exiting} \cite{teerapittayanon2016branchynet}. In this case, the forward pass is accelerated by enabling the model to yield a prediction (token) at intermediate layers, rather than passing through all the layers as is traditionally done.

A key component of early-exit models is the \emph{confidence score}, computed at every candidate exit, which determines whether the current prediction is of sufficient quality to terminate the forward pass and return the prediction. While various confidence measures have been proposed, most are derived from the predictive distribution at the given exit (e.g., maximum softmax probability). However, this poses a problem when applying early-exiting to LLMs \citep{elbayad2019depth, schuster2022confident, bae2023fast, varshney2023accelerating}, where obtaining the predictive distribution requires mapping the current hidden representation to the vector of logits over all possible tokens. Given the large vocabulary sizes used in modern LLMs ($\approx30$-$256\textrm{K}$) \cite{villalobos2024run, tao2024scaling}, such confidence estimation introduces significant computational overhead. This is one of the
main reasons behind the previously observed paradox, where early-exiting in LLMs resulted in less efficient inference compared to standard, non-accelerated models (both in terms of FLOPs \cite{schuster2022confident} and latency \cite{bae2023fast}), thereby defeating its original purpose.

In this work, we improve the efficiency of confidence estimation in early-exit LLMs. Specifically, we propose to map the hidden representation of the model to the full vocabulary only at the first couple of initial candidate exits, and use the resulting predictive distribution to identify the top $K$ most likely tokens. We then prune the weight matrix (which maps hidden representations to logits over tokens) based on the most likely tokens found and use the pruned weights at all subsequent candidate exits (\cref{fig:softmax_met}). Our design is motivated by the empirical observation that the token predicted at the final layer is among the top tokens already in the early layers of the forward pass (\cref{fig:pruning_final}). In our experiments, we demonstrate that \emph{dynamic vocabulary pruning} improves the FLOPs and time efficiency of confidence estimation in early-exit LLMs while preserving competitive performance. Importantly, our design is lightweight, as it is entirely post-hoc and requires no finetuning or the introduction of new model parameters

\section{Preliminaries}
\label{sec:prelim}

Let $\mathcal{Y}$ denote the vocabulary (or token) space, with size $|\mathcal{Y}| = d_{\text{vocab}}$. Further, for $\: x_i \in \mathcal{Y}$, let $(x_1, \ldots, x_t)$  represent the input sequence, comprising both the tokens in the prompt and those generated upon time $t$ by the model.

\paragraph{Autoregressive Decoding in LLMs} To predict the next token in the sequence, most modern language models employ the transformer architecture \cite{vaswani2017attention}. In a transformer model, the input sequence is passed through $L$ layers, each consisting of a multi-head attention and a feed-forward block, yielding a sequence of hidden representations $\{ \mathbf{h}_t^{\ell}\}_{\ell=1}^L, \mathbf{h}_t^{\ell} \in \mathbb{R}^{d_{\text{model}}}$. After processing through all layers, the final next token distribution is obtained via 
\[
p\left(x_{t+1} | \mathbf{h}_t^L\right) = \text{softmax}(\mathbf{W} \mathbf{h}_t^L) \: .
\]
$\mathbf{W} \in \mathbb{R}^{d_{\text{vocab}} \times d_{\text{model}}}$ is a weight matrix, also referred to as the \emph{unembedding} matrix, that projects the final hidden state $\mathbf{h}^L_t$ back to the token space $\mathcal{Y}$. The newly predicted token $x_{t+1}$ is then added to the input sequence, and the (autoregressive) generation process is repeated until termination.

\paragraph{Early-Exiting in LLMs}  Observe how decoding in LLMs, as introduced above, requires passing through all L layers for every token in the generated sequence, resulting in a slow inference process. To mitigate this, early-exiting (EE) mechanisms have been proposed \cite{elbayad2019depth, schuster2022confident}, allowing the model to predict tokens at intermediate layers if sufficiently confident. Specifically, for each layer $\ell$, a confidence score $c_t^\ell \in [0,1]$ and an exiting threshold $\lambda_t^\ell \in [0,1]$ are defined. The early prediction is returned as soon as the confidence at the current layer exceeds the threshold:

\begin{equation}
x_{t+1} := \begin{cases}
\arg\max p\left(x_{t+1} | \mathbf{h}_t^1\right) & \text{if } c_t^1 \geq \lambda_t^1, \\
\arg\max p\left(x_{t+1} | \mathbf{h}_t^2\right) & \text{if } c_t^2 \geq \lambda_t^2, \\
\vdots & \vdots \\
\arg\max p\left(x_{t+1} | \mathbf{h}_t^L\right) & \text{otherwise}.
\end{cases}
\end{equation}
Note that it is common to reuse the final weight matrix $\mathbf{W}$ at earlier exits \cite{schuster2022confident, elhoushi2024layer}, i.e.,  $p\left(x_{t+1} | \mathbf{h}_t^{\ell}\right) = \text{softmax}(\mathbf{W} \mathbf{h}_t^{\ell}), \forall \ell = 1,\ldots, L$, which avoids instantiating a separate unembedding matrix at each exit and prevents introducing a significant number of additional model parameters. Moreover, for simplicity, it is common to assume a fixed and shared threshold $\lambda$ across all exits and tokens \cite{jazbec2024fast}.
\section{Dynamic Vocabulary Pruning}
\label{sec:dvp}
\paragraph{Softmax Based Confidence Measures} As introduced in \cref{sec:prelim}, a confidence measure is necessary to determine whether the model's current prediction is of sufficient quality to terminate the forward pass and return an early prediction. Most commonly, the so-called softmax based measures are used, e.g. the maximum softmax probability $c_t^{\ell} = \max p(x_{t+1} | \mathbf{h}_t^{\ell})$. 
However, this requires computations involving the full unembedding matrix $\mathbf{W}$ at every exit, which is expensive due to the large $d_{\text{model}}$ and $d_{\text{vocab}}$ used in modern LLMs.\footnote{Confidence measures based directly on the hidden states $\mathbf{h}_t^{\ell}$ have also been explored, but they have been shown to result in slower exiting compared to softmax-based scores \cite{schuster2022confident}.}  While this may be less concerning for latency—since the execution of the next transformer block can proceed in parallel with the confidence estimation—it still reduces the overall efficiency of the forward pass. For example, in CALM \cite{schuster2022confident}, the authors report that their early-exit model with softmax confidence is approximately twice as expensive in terms of FLOPs compared to a static model (i.e., without early exiting), despite requiring around 50\% fewer layers per token on average (see Table 2 in \cite{schuster2022confident}). This can make early-exiting impractical, especially in scenarios where FLOPs are a critical constraint (e.g., on device).
\begin{figure}[t]
    \centering
    \includegraphics[width=\textwidth]{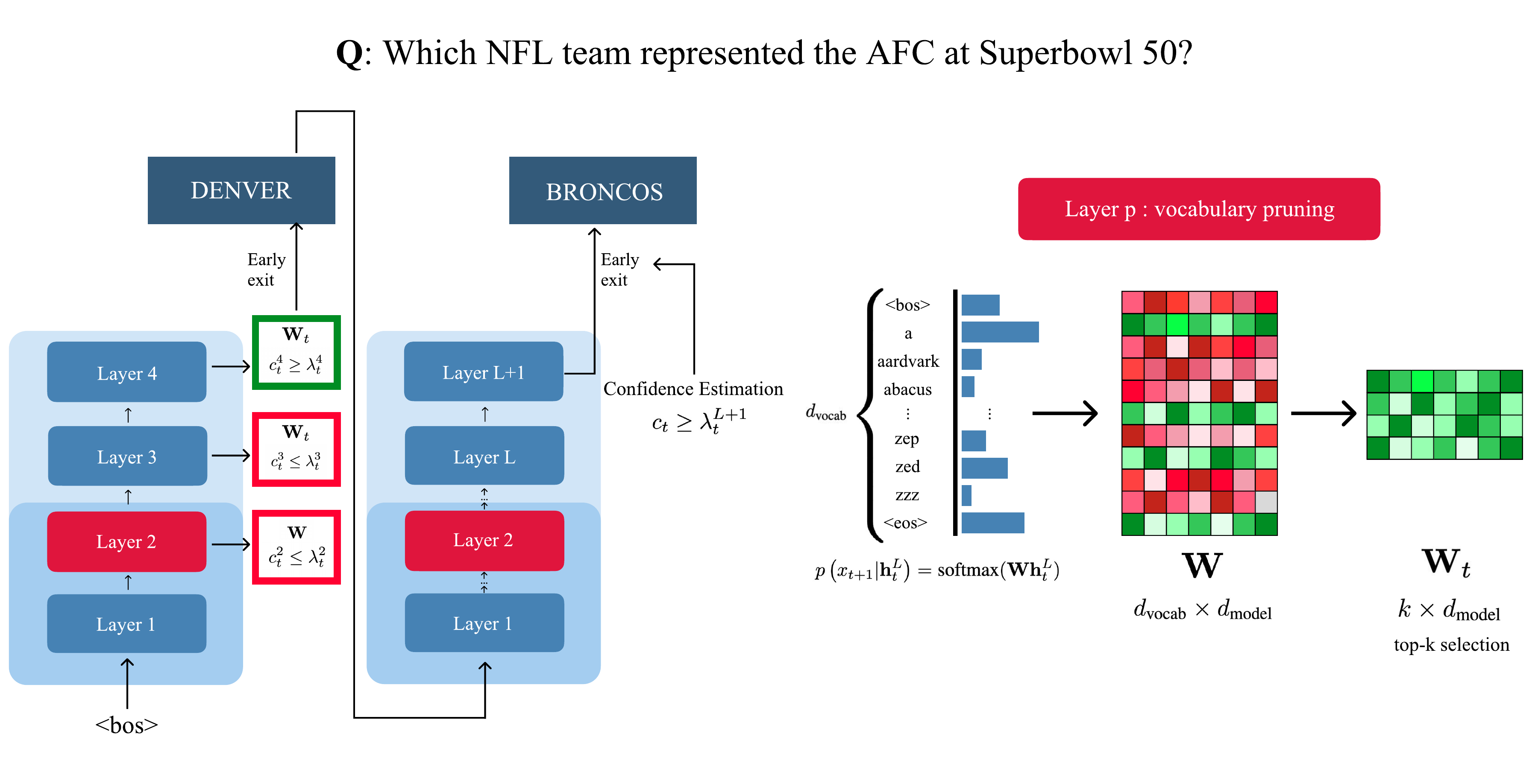}
    \caption{\textit{Left}: Illustration of our vocabulary pruning setup in Transformer models during inference. The model evaluates the input question with an Early Exiting objective where the vocabulary is reduced at a fixed layer $p = 2$ in the reference figure. At each layer $\ell$, the model computes a confidence estimation $c_t^\ell$ and compares it against a threshold $\lambda_t^\ell$. When the model achieves sufficient confidence about the token to predict at layer $\ell+1$, the token is returned. \textit{Right}: Visualization of our proposed pruning mechanism. At exit $p$, we first identify the top $K$ most likely tokens, which are used to subsample the rows of the unembedding matrix $\mathbf{W}$. The resulting pruned matrix $\mathbf{W}_t$ is then used for confidence estimation at all subsequent exits.}
    \label{fig:softmax_met}
\end{figure}

\begin{wrapfigure}{R}{0.5\textwidth}
  \centering
  \includegraphics[width=0.48\textwidth]{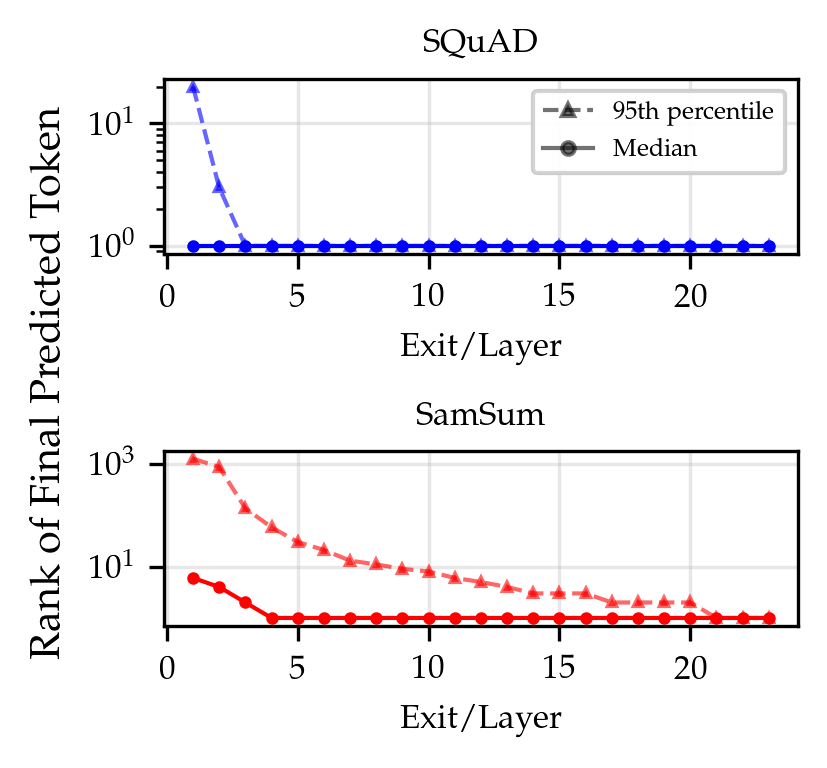}
  \caption{Rank (log-scale) of the final predicted token across model exits/layers on SQuAD \cite{rajpurkar2016squad} and SamSum \cite{gliwa2019samsum} using the early-exit version of the \texttt{T5-large} model \cite{bae2023fast}. We observe a clear trend of very early layers showing a low average rank for the final predicted tokens, which motivates our dynamic vocabulary pruning approach.}
  \label{fig:pruning_final}
  \vspace{1\baselineskip}
\end{wrapfigure}

\paragraph{Dynamic Vocabulary Pruning} To reduce the overhead of confidence estimation in early-exit LLMs, we investigate whether the full computation with $\mathbf{W}$ is indeed necessary at every candidate exit. In particular, we study how quickly the token predicted after passing through all the layers appears among the most likely tokens at earlier layers. As depicted in \cref{fig:pruning_final}
, we note that this occurs quite early in the forward pass. For example, in the case of the CALM model \cite{schuster2022confident} on the SQuAD dataset, we observe that the token predicted at the last layer appears among the top $10$ most likely tokens already at the 2nd layer in $95\%$ of cases.\footnote{We find that early-exit finetuning (see Eq. (6) in \cite{schuster2022confident}) is important for ensuring faster token convergence. See Appendix \ref{app:ee-ft} for more details.} This suggests that mapping to the full vocabulary becomes redundant after a certain (early) layer.

We make use of this empirical observation in the design of our pruning solution. Specifically, we propose to map the hidden states to the full vocabulary only up to and including exit $p$ (e.g., $p=1$ or $p=2$). Then, we use the logits vector $\mathbf{l}_t^{p} = \mathbf{W} \mathbf{h}_t^{p} \in \mathbb{R}^{d_{\text{vocab}}}$ to identify the top $K$ most likely tokens and use those to \emph{prune} the embedding matrix $\mathbf{W}$ (by selecting the rows associated with the indices of the most likely tokens, see \cref{fig:softmax_met}
). We denote the pruned matrix as $\mathbf{W}_t \in \mathbb{R}^{K \times d_{\text{model}}}$ and use it to compute the confidence at all subsequent layers. The index $t$ in $\mathbf{W}_t$ highlights the dynamic nature of our pruning, i.e., it is performed independently for each token in the generated sequence. Since $K \ll d_{\text{vocab}}$, the cost of confidence estimation is significantly reduced.

To determine the optimal pruning hyperparameters ($p$ and $K$), we suggest using a small calibration dataset and finding the smallest values for which the performance drop remains negligible. We leave the incorporation of more principled selection mechanisms \cite{jazbec2024fast} for future work. 


\newpage
\section{Experiments}
\label{sec:exp}

We closely follow the experimental setup of \citet{schuster2022confident}. Specifically, we use the \texttt{T5-large} model \cite{2020t5} and consider the tasks of question answering (SQuAD \cite{rajpurkar2016squad}) and text summarization (SamSum \cite{gliwa2019samsum}). As a baseline, we use the CALM model \cite{schuster2022confident} with (full) softmax confidence estimation. Our code is publicly available\footnote{\url{https://github.com/MatteoNulli/Vocabulary_pruning/tree/main}} and we provide further implementation details in \cref{app:impl}.

The results are presented in \cref{tab:comparison}. First, we observe that for both tasks, our proposed Dynamic Vocabulary Pruning (DVP) either matches the baseline or incurs only a negligible performance drop. Under the same conditions, it outperforms the softmax implementation in CALM \cite{schuster2022confident}, in terms of FLOPs and time required for exit decisions. For instance, on the SQuAD dataset, using a conservative exit threshold ($\lambda = 0.99$), our DVP achieves the same F1 score ($90.6$) while requiring $\sim7\mathrm{x}$ fewer FLOPs than the full softmax baseline. Importantly, unlike other FLOP-efficient confidence measures (e.g., hidden state saturation from \cite{schuster2022confident}), our DVP does not require evaluating additional blocks/layers, as evidenced by the similar average exit block indices compared to the baseline. This observation confirms that the pruned vocabulary terms are indeed the ones not usually predicted by the model. Moreover, while not
the primary focus of our work, it is encouraging that DVP also results in reduced latency (i.e., shorter time required to compute exit confidence). Overall, these results suggest that our dynamic vocabulary pruning method effectively addresses the high cost of confidence estimation in early-exit LLMs with little to no impact on overall performance.

\begin{table}[ht]

    \centering

    \resizebox{\textwidth}{!}{%

    \begin{tabular}{llcccccc}

        \toprule

        \textbf{Dataset} & \textbf{Conf. $\lambda$} & \textbf{Method} & \textbf{Score  ($\uparrow$)} & \textbf{FLOPs/Token ($\downarrow$)} & \textbf{Avg. Exit ($\downarrow$)} & \textbf{Conf. Time (s) ($\downarrow$)} \\

        \midrule

        \multirow{4}{*}{\centering \textbf{SQuAD} \citep{rajpurkar2016squad}} & \multirow{2}{*}{$0.6$} & CALM & \textbf{87.5} & 2.21 $\times$ $10^{8}$  & \underline{2.4} & 44.5 \\

        & & + DVP (ours) & 87.4 & \textbf{1.97 $\times$ $\mathbf{10^{8}}$} & \underline{2.4} & \textbf{40.8}  \\

        \\[-0.20cm] 

        & \multirow{2}{*}{$\approx 0.99$} & CALM & \underline{90.6} & 13.91 $\times$ $10^{8}$  & 20.9 & 499.9\\

        & & + DVP (ours) & \underline{90.6} & \textbf{1.99 $\times$ $\mathbf{10^{8}}$}  & \textbf{20.8} & \textbf{413.1} \\

        \midrule

        \multirow{4}{*}{\centering \textbf{SamSum} \citep{gliwa2019samsum}} & \multirow{2}{*}{$0.6$} & CALM & \textbf{33.8} & 4.21 $\times$ $10^{8}$ & 5.5  & 90.0 \\

        & & + DVP (ours) & 33.7 & \textbf{2.01 $\times$ $\mathbf{10^{8}}$}  & \textbf{5.4} & \textbf{81.0} \\

        \\[-0.20cm] 

        & \multirow{2}{*}{$\approx 0.99$} & CALM & \underline{43.1} & 11.13 $\times$ $10^{8}$ & 16.5 & 162.0 \\

        & & + DVP (ours) & \underline{43.1} & \textbf{2.12 $\times$ $\mathbf{10^{8}}$}  & \textbf{16.4} & \textbf{136.0} \\

        \bottomrule

    \end{tabular}

    }

    \vspace{0.2cm}

    \caption{Summary of efficiency gains for our dynamic vocabulary pruning (DVP) compared to CALM \cite{schuster2022confident} for two different exiting thresholds $\lambda$ (0.6 and $\approx 0.99$). 
    To measure the performance quality, we report F1 score for SQuAD \citep{rajpurkar2016squad} and Rouge-L metric for SamSum \citep{gliwa2019samsum}. Additionally, we outline the amount of FLOPs per generated token and average early-exit layer across generated tokens (note that the full \texttt{T5-large} model has 24 layers). We also report the total time spent on confidence estimation for the entire test set.}
    \label{tab:comparison}
    \vspace{-2\baselineskip}

\end{table}

\section{Conclusion \& Future Work}
Our work tackles the high cost of confidence estimation in early-exit LLMs, which arises from large vocabulary sizes. By dynamically pruning the vocabulary for every generated token, we demonstrate that efficient confidence computation is achievable without compromising performance. Our proposed vocabulary pruning is completely post-hoc, making it nicely compatible with existing pretrained early-exit LLMs. We hope our findings encourage a reconsideration of the trend towards sacrificing model adaptivity (i.e., reducing the number of possible exits \cite{bae2023fast}) due to the growing computational cost of exiting decisions. In future work, it would be valuable to validate our approach on other early-exit LLMs \cite{varshney2023accelerating} and explore more advanced pruning mechanisms (e.g., using product-of-experts ensembles across exits \cite{jazbec2024towards}) beyond the simple top-K strategy used here. Future work could also  investigate the impact of dynamic vocabulary pruning on confidence calibration \cite{meronen2024fixing}.

\setcitestyle{numbers}
\bibliography{neurips_2024}
\bibliographystyle{abbrvnat}

\newpage


\appendix
\section*{Appendix}

\section{Implementation Details}
\label{app:impl}
We ran our experiments on 1x Nvidia A100 80GB - SMX4 GPU. Our code is available at \url{https://github.com/MatteoNulli/Vocabulary_pruning/tree/main}.

We report all the relevant early-exiting hyperparameters for our experiments in Table \ref{table:hyperparams}. Our DVP approach introduces $p$ and $K$ which represent the pruning exit index and the pruned vocabulary size, respectively. The \texttt{top-2 diff} strategy indicates that the exit confidence $c_{t}^{\ell}$ is computed as the difference between the probabilities of the top two tokens. The \texttt{decaying} threshold $\lambda_t$ means that the exit threshold decreases for later tokens in the generated response (see Eq. (5) in \cite{schuster2022confident}).

\begin{table}[ht]
\begin{center}
\scalebox{0.95}{
\begin{tabular}{l|c c c c}
\toprule
 & \multicolumn{2}{c}{\textbf{SQuAD}} & \multicolumn{2}{c}{\textbf{SamSum}} \\
                  & CALM & DVP & CALM & DVP \\
\midrule
$p$                & -    & 2   & -    & 2   \\
$K$                & -    & 64  & -    & 512 \\
$c_t^{\ell}$ & \texttt{top-2 diff} & \texttt{top-2 diff} & \texttt{top-2 diff} & \texttt{top-2 diff} \\
$\lambda_t$    & \texttt{static} & \texttt{static} & \texttt{decaying} ($\tau = 4$) & \texttt{decaying} ($\tau = 4$) \\
\bottomrule
\end{tabular}
}
\end{center}
\caption{Main early-exit hyperparameters used in our experiments.}
\label{table:hyperparams}
\end{table}






\section{Additional Experiments}
\label{app:ee-ft}

In Section \ref{sec:dvp}, we reported that, for an early-exit LLM like CALM \cite{schuster2022confident}, the token predicted at the final layer is often among the top-K predicted tokens quite early in the process. Here, we investigate the effect of adapting the unembedding matrix $\mathbf{W}$ to intermediate representations $\mathbf{h}_t^{\ell}$ through early-exit finetuning (see Eq. (6) in \cite{schuster2022confident}). The results, displayed in Figure 3, show that the T5 model \cite{2020t5} without early-exit finetuning exhibits slower convergence compared to the T5 model that has undergone early-exit finetuning (which corresponds to the CALM model). This finding is important for our dynamic vocabulary pruning proposal, as faster convergence enables the selection of lower values for pruning parameters ($p$ and $K$), resulting in larger efficiency savings.

\begin{figure}[h]
  \centering
  \includegraphics[width=\textwidth]{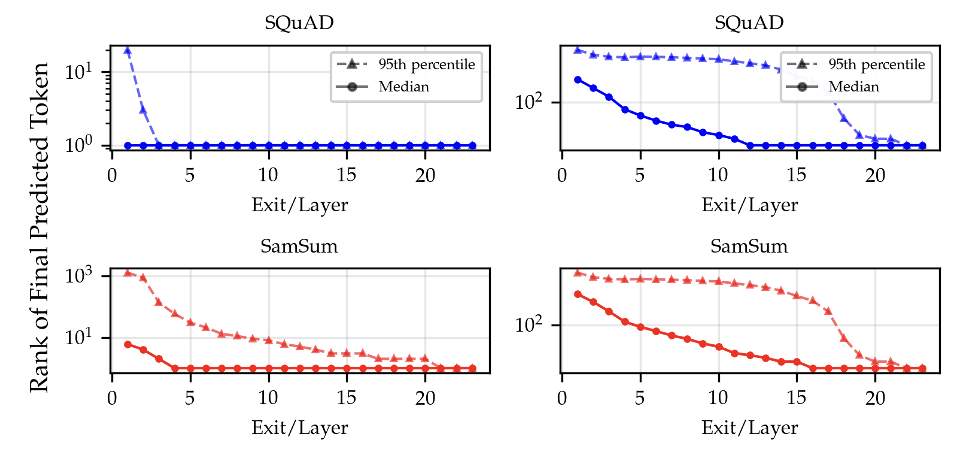}
  \caption{Rank (log-scale) of the final predicted token across model exits/layers on SQuAD \cite{rajpurkar2016squad} and SamSum \cite{gliwa2019samsum}.
\textit{Left:} Results based on CALM \cite{schuster2022confident}, the early-exit version of the \texttt{T5-large} model \cite{bae2023fast}. These are the same results as those shown in Figure \ref{fig:pruning_final}, included here for easier comparison.
\textit{Right:} Results based on the \texttt{T5-large} model \cite{2020t5}, where the non-adapted original unembedding matrix is used at intermediate layers to facilitate early-exiting.}
  \label{fig:pruning_final-non-ft}
  \vspace{1\baselineskip}
\end{figure}

\end{document}